\def\year{2022}\relax
\documentclass[letterpaper]{article} 
\usepackage{aaai22}  
\usepackage{times}  
\usepackage{helvet}  
\usepackage{courier}  
\usepackage[hyphens]{url}  
\usepackage{graphicx} 
\usepackage{natbib}  
\usepackage{caption} 
\DeclareCaptionStyle{ruled}{labelfont=normalfont,labelsep=colon,strut=off} 
\usepackage{epsfig}
\usepackage{graphicx}
\usepackage{amsmath}
\usepackage{amssymb}
\usepackage{booktabs} 
\usepackage{color}
\usepackage{indentfirst}
\usepackage{multirow}
\usepackage{xcolor}
\usepackage{algorithm,algpseudocode}
\newcommand{\heading}[1]{\noindent\textbf{#1}}

\usepackage{newfloat}
\usepackage{listings}
\usepackage{algpseudocode}
\usepackage{amsmath}  
\usepackage{graphics}
\usepackage{newfloat}
\usepackage{listings}
\floatstyle{ruled}
\newfloat{listing}{tb}{lst}{}
\floatname{listing}{Listing}
\title{MixSiam: A Mixture-based Approach to Self-supervised \\ Representation Learning}

\author{Xiaoyang Guo$^{1}$\thanks{The first two authors have equal contribution.} , Tianhao Zhao$^{1}$, Yutian Lin$^{1}$, Bo Du$^{1}$}
\affiliations{
$^{1}$School of Computer Science, Wuhan University\\ 
soulredeem@whu.edu.cn, happytianhao@whu.edu.cn\\
{yutianlin477@gmail.com, dubo@whu.edu.cn}

}

\usepackage{bibentry}

\begin{document}

\maketitle

\begin{abstract}
Recently contrastive learning has shown significant progress in learning visual representations from unlabeled data. The core idea is training the backbone to be invariant to different augmentations of an instance. While most methods only maximize the feature similarity between two augmented data, we further generate more challenging training samples and force the model to keep predicting discriminative representation on these hard samples. In this paper, we propose MixSiam, a mixture-based approach upon the traditional siamese network. On the one hand, we input two augmented images of an instance to the backbone and obtain the discriminative representation by performing an element-wise maximum of two features. On the other hand, we take the mixture of these augmented images as input, and expect the model prediction to be close to the discriminative representation. In this way, the model could access more variant data samples of an instance and keep predicting invariant discriminative representations for them. Thus the learned model is more robust compared to previous contrastive learning methods. Extensive experiments on large-scale datasets show that MixSiam steadily improves the baseline and achieves competitive results with state-of-the-art methods. Our code will be released soon.

\end{abstract}
\section{Introduction}
Learning discriminative image representation in an unsupervised/ self-supervised manner has attracted increasing interest~\cite{agrawal2015learning,doersch2015unsupervised,xie2021propagate}, for it gets rid of the costly manually-labeled data and achieves promising performance on many down-stream tasks~\cite{larsson2019fine,hung2019scops,doersch2017multi}. These methods generally design pretext tasks and learn the representation from the label generated by the tasks, such as rotation predicting~\cite{komodakis2018unsupervised}, jigsaw~\cite{noroozi2016unsupervised,kim2018learning}, in-painting~\cite{pathak2016context}, colorization~\cite{zhang2016colorful,larsson2017colorization} and clustering~\cite{noroozi2018boosting,caron2018deep}.
Among them, many state-of-the-art methods employ the principle of contrastive learning~\cite{NEURIPS2020_f3ada80d,chen2020simple,chen2021exploring,he2020momentum} and achieve remarkable progress. 

\begin{figure}[ht]
   \centering
    \includegraphics[width=0.9\linewidth]{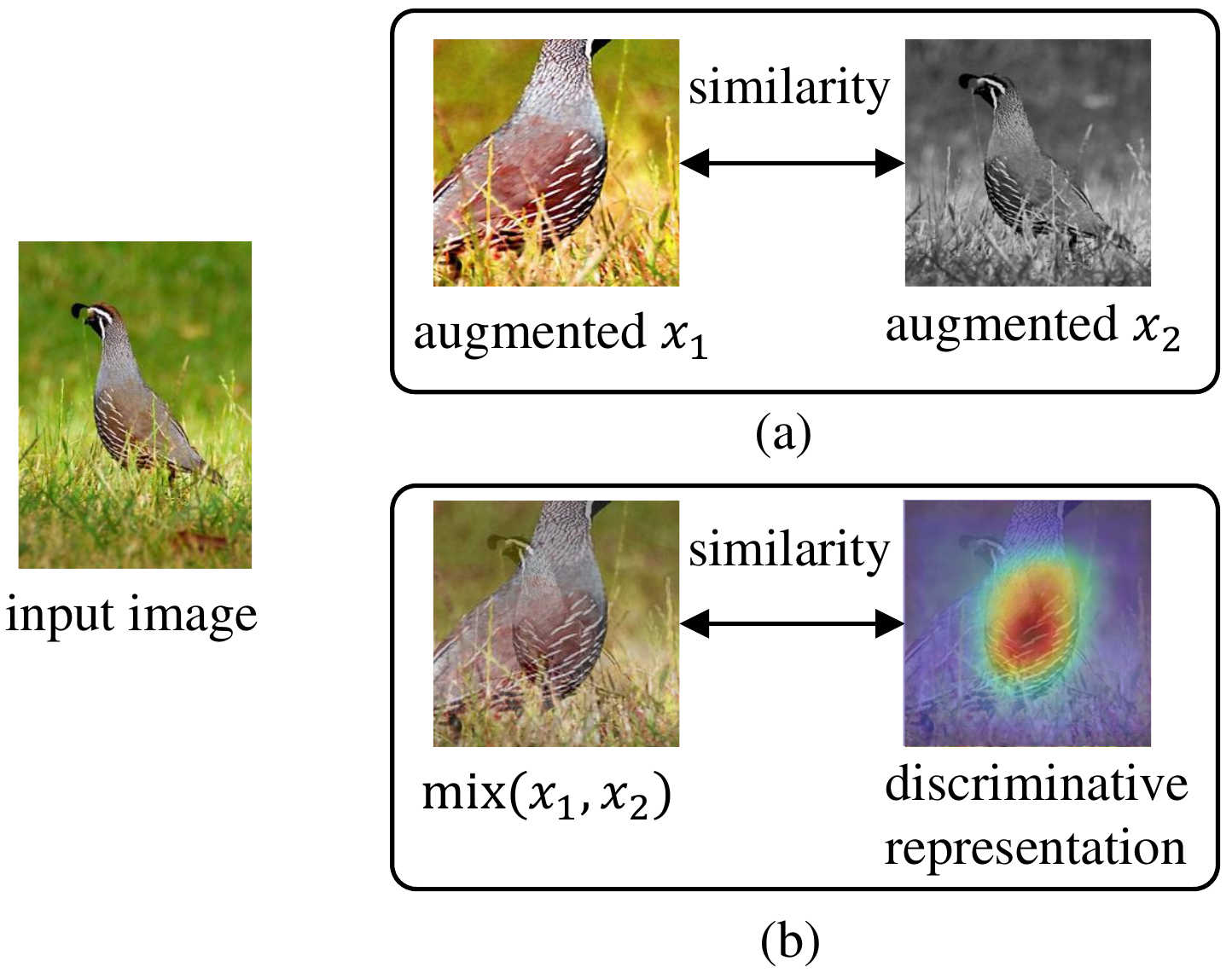}
   \caption{Constructed pairs of contrastive learning. (a) In most previous works, the similarity between augmented views is minimized. (b) The feature of the hard sample is supposed to share the same representation with the discriminative representation.}
   \label{fig:motivation}
\end{figure}

Contrastive learning generally constructs a pretext task to discriminate from augmented positive or negative image pairs, whereas the representations of different augmented views of the same instance are supposed to be close, while the representation of different instances is supposed to be far away. 
A branch of methods utilize both of the positive and negative pairs and carefully treat negative samples with large batch size~\cite{oord2018representation,chen2020simple} or memory bank~\cite{he2020momentum,wu2018unsupervised}. While in some recent works, such as BYOL~\cite{NEURIPS2020_f3ada80d} and SimSiam~\cite{chen2021exploring}, a siamese network is proposed, where only positive pairs are used to train a siamese network and maintain competitive performance. As shown in Fig.~\ref{fig:motivation} (a), previous contrastive learning methods with the siamese structure only maximize the similarity between positive feature pairs. However, the two augments have large mutual information, which might decrease the ability of the model to discriminate images with large intra-class variations. We argue that learning from unseen samples and the discriminative representations could contribute to siamese contrastive, as shown in Fig.~\ref{fig:motivation} (b). By considering both of the relationships, the robustness and the discrimination ability of learned representations are improved.

In this paper, we propose to learn from both the augmented easy positive pairs and the hard mixed images. Inspired by MixUp~\cite{zhang2018mixup}, two augmented views of one instance are linearly mixed to generate the hard training sample, which extends the training distribution and helps to prevent the neural network from overfitting. On the other hand, the two augmented views are fed to the backbone to obtain the augmented features, which are further aggregated with element-wise maximum. By conducting the maximum operation, the most discriminative information from the two features is reserved, where the aggregated feature is denoted as the discriminative representation. In addition to the siamese contrastive learning framework, we propose to maximize the similarity between the feature of the mixed image and the discriminative representation. We expect the model to predict invariant discriminative representations on these generated challenging samples.

The structure of MixSiam is shown in Fig.~\ref{fig:structure} (b). 
The two augmented images of the original instance and their mixed image are fed into the encoder network, where the three encoders directly share weights with each other. The output of the predictor network in the right is forced to share the representation with the discriminative representation obtained by the two encoder networks in the left. The strong similarity constraint helps to extract robust features with discrimination ability. The experiments on various datasets demonstrate that the proposed method can consistently boost the performance over the SimSiam~\cite{chen2021exploring} baseline by learning from the mixed images and the discriminative representations. 

Our contribution can be summarized in three-fold:

(1) We propose MixSiam, a mixture-based approach upon the traditional siamese contrastive, which learns invariant discriminative representations from the generated hard mixed images. With these hard training samples and the strong constraint of the discriminative representation, a more robust model is learned. 

(2) We ablate our approach to discuss the impact of different feature aggregation methods, and prove the effectiveness of obtaining the most discriminative representation, which could also inspire following research.

(3) Our method achieves a stronger backbone network, which has been effectively verified on the standard ImageNet and multiple small-scale datasets such as CIFAR10, CIFAR100, and Food101. With a much smaller batch size training, MixSiam achieves competitive classification accuracy than existing state-of-the-art methods.

\section{Related Work}
\subsection{Pretext Tasks}
Most self-supervised learning methods are based on pretext tasks, which train networks with task-defined labels and learn visual representations through the process. A branch of methods is based on image generation, which extracts invariant features of images by image inpainting~\cite{pathak2016context}, image colorization~\cite{zhang2016colorful,larsson2017colorization} or image generation with GAN~\cite{radford2015unsupervised,donahue2017adversarial}. Another category of the methods is based on the semantic context of the images~\cite{noroozi2016unsupervised,komodakis2018unsupervised}. For example, in~\cite{noroozi2016unsupervised}, the network is trained to solve the image jigsaw puzzle. In~\cite{komodakis2018unsupervised}, the network is trained to recognize the geometric transformations. During these pretext tasks, the networks learn to discriminate the semantic context, and provide semantic-aware representations. In addition, some clustering-based methods~\cite{noroozi2018boosting,caron2018deep} are also based on the context similarity. These methods are usually designed for specific domains, while the generality is limited.

\subsection{Contrastive Learning}
In recent years, the rapid development of contrastive learning~\cite{hadsell2006dimensionality,wu2018unsupervised,tian2020contrastive,zhuang2019local,misra2020self} has brought a greater breakthrough to self-supervised learning. The main idea of contrastive learning is to pull the positive pairs closer, while pushing the negative pairs away from each other in the embedding space. Previous works demonstrate that contrastive learning benefits from observing sufficient negative samples~\cite{jaiswal2021survey}. Follow this principle, InstDist~\cite{wu2018unsupervised} builds a memory bank storing all representations of instances, which increases the size of negative pairs. In MoCo~\cite{he2020momentum}, a momentum update mechanism is adopted to maintain a queue of negatives to build a stable data distribution with a large number of negative samples. Following this work, in SimCLR~\cite{chen2020simple}, large-scale batch size is adopted to provide negative samples instead of the memory bank. In recent, some studies~\cite{NEURIPS2020_f3ada80d,chen2021exploring} achieve promising performance with only positive pairs. BYOL~\cite{NEURIPS2020_f3ada80d} adopts a siamese network where one branch is a momentum encoder. It directly predicts the output of one view from another view. SimSiam~\cite{chen2021exploring} further improved by adopting a stop gradient that iteratively receives gradient information parameters to prevent the siamese network from collapsing. 
In this paper, we make an attempt beyond the original siamese framework, where another branch with the mixed harder sample as input is utilized to predict invariant discriminative representation. 

\begin{figure*}[ht]
   \centering
    \includegraphics[width=0.9\linewidth]{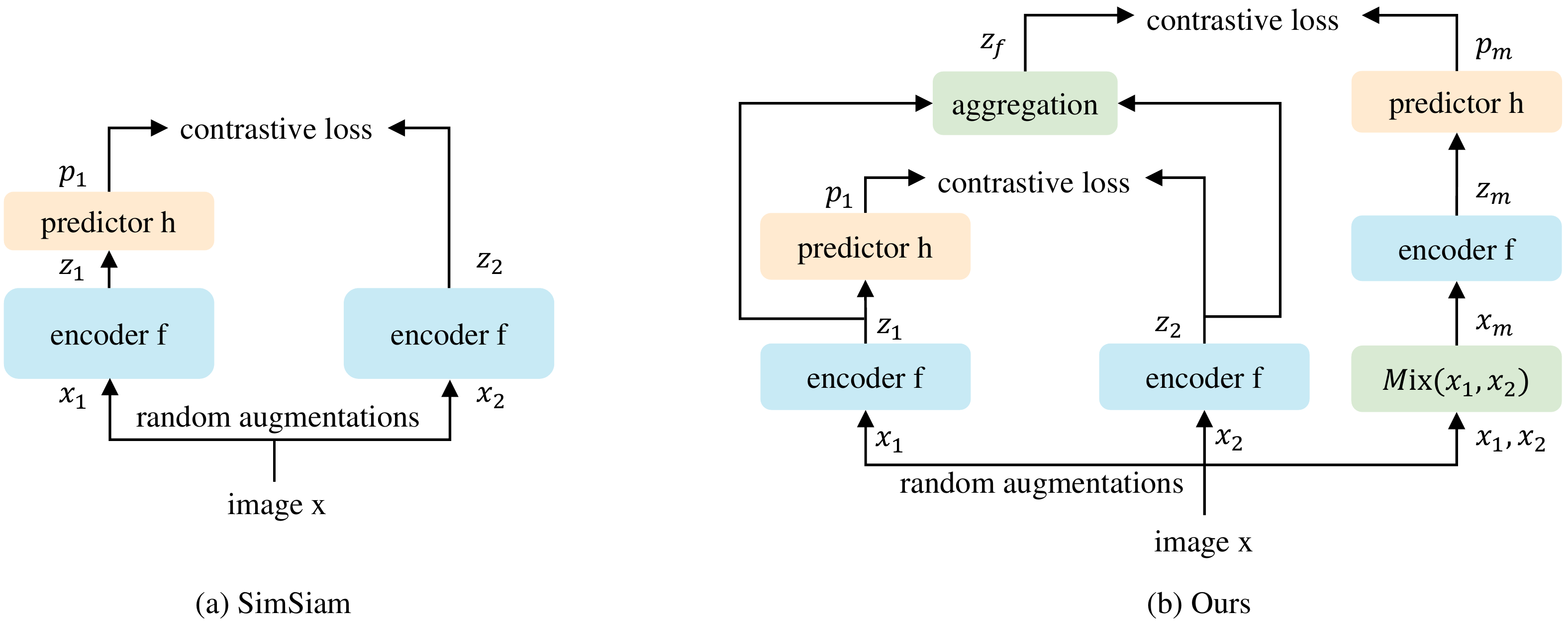}
   \caption{The illustration of two different contrastive architectures. (a) SimSiam architecture. Two augmented views of one image are processed by the same encoder network. Then a prediction MLP is applied on one side, and a stop-gradient operation is applied on the other side. The model maximizes the similarity between both sides. (b) Our architecture. Two augmented views, as well as a sample of their mixture, are fed into the network. Except for the original structure, a prediction MLP is applied to obtain the output embedding of the mixed image, and an aggregation operation is applied to the other two sides with a stop-gradient operation. The final loss maximizes the similarity between the embedding of the image mixture and the aggregated embedding.}
   \label{fig:structure}
\end{figure*}

\subsection{Image Mixture Augmentation}
As an image augmentation strategy, MixUp~\cite{zhang2018mixup} and its variants~\cite{verma2019manifold,kim2020puzzle,yun2019cutmix} are proposed to improve generalization and alleviate adversarial perturbation and shown effectiveness on both supervised~\cite{verma2019manifold,guo2019mixup} and semi-supervised~\cite{lamb2019interpolated,berthelot2019mixmatch} settings. Manifold Mixup~\cite{verma2019manifold} is proposed to use latent representation mixture to create virtual samples for supervised learning. In MoCHi~\cite{kalantidis2020hard}, hard negative samples are mixed on the embedding space to improve the generalization of the visual representations. I-Mix~\cite{lee2021mix} introduces virtual labels in a batch and mix data instances and their corresponding virtual labels in the input and label spaces, respectively. 
Similarly, MixCo~\cite{kim2020mixco} adopts a mix-up training strategy in contrastive learning to build semi-positive samples. 
Un-Mix~\cite{shen2020mix} randomly mixes all the samples in a mini-batch with the smooth label to smooth decision boundaries and prevent the learner from becoming over-confident. 
In this paper, we adopt image mixture in the image space to obtain the hard mixed images for training. Instead of directly predicting the mixed images to the mixed labels, we further force the representation of the mixed images maintaining the most discriminative representation.

\section{Method}
As shown in Fig.~\ref{fig:structure} (a), previous SimSiam~\cite{chen2021exploring} adopts a siamese network directly predicting the output of one view from another view, where one branch adopts a stop gradient to prevent from collapsing. As shown in Fig.~\ref{fig:structure} (b), our framework contains two kinds of positive pairs with three branches for similarity optimization. Comparing to the previous siamese structure, we further adopt a branch with mixed images, which is forced to predict invariant discriminative representations for these hard samples. 
In the following, we first revisit SimSiam~\cite{chen2021exploring} as one of the leading methods in this domain. Next, we introduce our approach in detail. 
\subsection{Contrastive Learning with Siamese Network}
SimSiam~\cite{chen2021exploring} proposes to directly measure the similarity between positive pairs without construct negative pairs. As shown in Fig.~\ref{fig:structure} (a), two random augmented views of the same instance are adopted as the positive pair. These two views are then fed through an encoder to obtain the positive embedding pairs for contrastive learning. 

Formally, for each training instance $x$, two random augmented views $x_1$ and $x_2$ are generated and fed through the encoder $f$. The encoder $f$ is typically a ResNet-50 with a projection MLP head, which shares weight between the two branches. We denote the prediction MLP head as $h$, which transforms the output of one view to match the other view. The two output embeddings are then defined as $p_1=h(f(x_1))$ and $z_2=f(x_2)$. Then the distance (\textit(i.e.,) negative cosine similarity) between two views is calculated as follows:
\begin{equation}
\mathcal{D}(p_1,z_2)=-\frac{p_1}{\vert\vert p_1\vert\vert_2}\cdot \frac{z_2}{\vert\vert z_2\vert\vert_2,}
\end{equation}
where $\vert\vert \cdot \vert\vert$ is $l_2$-norm. Then the distance between positive pairs is minimized, where the loss function is defined as:
\begin{equation}
\mathcal{L}=\frac{1}{2}\mathcal{D}(p_1,z_2)+\frac{1}{2}\mathcal{D}(p_2,z_1).
\label{eq:loss1}
\end{equation}

As an important component, a stop-gradient ($stopgrad(\cdot)$) operation is adopted to the encoder to prevent from collapsing. Then the equation~\ref{eq:loss1} is re-formulated as:
\begin{equation}
\mathcal{L}=\frac{1}{2}\mathcal{D}(p_1,stopgrad(z_2))+\frac{1}{2}\mathcal{D}(p_2,stopgrad(z_1)).
\end{equation}

Since SimSiam solely relies on positive pairs generated by random augmentation of the same instance, the mutual information between image pairs is large, which might decrease its capacity to be invariant to images with large intra-class variations. We try to address this issue by introducing our method.
\subsection{Mix-based Feature Learning Strategy}
In order to enrich the latent representation, we enlarge the support of the training distribution by introducing mixed virtual examples. These samples are hard and unnatural, which would boost the robustness of the model. Rather than directly training with the mixed images and the corresponding mixed labels as in Un-Mix~\cite{shen2020mix} or I-mix~\cite{lee2021mix}, we propose to give the hard samples a stronger constraint, where the model is supposed to predict invariant discriminative representations for the mixed hard samples. Through the process, more discriminative representations can be achieved.

For a given image $x$, SimSiam generates two randomly augmented images  $x_1$ and $x_2$ to form the positive pair. In addition, we define $x_m$ as the linear mixed image of $x_1$ and $x_2$, which is calculated as:
\begin{equation}
x_m=\lambda_{mix} x_1 + (1-\lambda_{mix}) x_2,
\label{eq:mix}
\end{equation}
where $\lambda_{mix} \in [0,1]$. In this way, the training distribution is extended by linear interpolations of image vectors. 

For any augmented image $x_k$, we denote $p_k \triangleq h\left(f\left(x_k\right)\right)$ and $z_k \triangleq f\left(x_k\right)$, where $f$ is the encoder network, $h$ is the predictor network and $k$ represents any arbitrary image index. Then, the output features can be defined as  $z_1=f(x_1)$, $z_2=f(x_2)$ and $p_m=h(f(x_m))$. 

To measure the similarity between two augmented views, we adopt the original symmetrized loss:
\begin{equation}
\mathcal{L}_{siam}=\frac{1}{2}\mathcal{D}(p_1,stopgrad(z_2))+\frac{1}{2}\mathcal{D}(p_2,stopgrad(z_1)).
\label{eq:losssiam}
\end{equation}

As for the mixed example $x_m$, a output feature $p_m=h(f(x_m))$ is obtained after fed through the encoder. Instead of predicting $x_m$ to the linear mixed vector of $z_1$ and $z_2$, we suppose the model to predict the most discriminate representation. Specifically, the output vector $z_1$ and $z_2$ are aggregated to produce an intermediate representation $z_f$ by the element-wise maximum function:
\begin{equation}
    z_f = maximum(z_1, z_2),
\end{equation}
where $z_f$ is adopted as the discriminative representation. Here, we adopt the element-wise maximum operation as the aggregation strategy, for it conducts feature selection, where the more discriminate features are maintained while the noise from the background is reduced. In the section of the experiment, multiple aggregation methods are investigated with experimental results, which indicates the superiority of element-wise maximum. 


The similarity between the mixed sample and the discriminative representation is then optimized as:
\begin{equation}
\mathcal{L}_{mix}=\mathcal{D}(p_m,stopgrad(z_f)).
\label{eq:lossmix}
\end{equation}

By optimizing the loss in Eq.~\ref{eq:lossmix}, the model is supposed to predict the discriminative representation $z_f$ to the feature $p_m$ of the mixed image. By employing Eq.~\ref{eq:mix}, the mixed image is more challenging with the target object in overlap and more noisy background. Eq.~\ref{eq:lossmix} acts like a strong constraint to force the model to learn from variant pairs, which further enhances the robustness of the model.

The final loss function is then calculated as follows:
\begin{equation}
\mathcal{L}=\lambda \mathcal{L}_{siam} + (1-\lambda) \mathcal{L}_{mix},
\label{eq:final}
\end{equation}
where $\lambda \in [0,1]$ balances the effect of easy positive pairs and the virtual mixed pairs.

In summary, based on the simple siamese network, our method enrich the latent representation by image mixup, and we further go beyond the simple positives by predicting the virtual example to the discriminative representation of the original instance. With the challenging virtual samples and the strong similarity constraint, the model learned to be invariant to the images with large intra-class variations. 
The whole procedure is summarized in Algorithm~\ref{algorithm}.

\begin{algorithm}[tb] 
  \begin{algorithmic}[1]  
    \Require Unlabeled training image $x$, encoder network $f$ with backbone and the projection mlp, prediction network $h$
    \Ensure
    Trained encoder $f$
    \State \textbf{Define:} 
    
    $aug(\cdot)$: function of random augmentation;
    
    $max(\cdot)$: function of element-wise maximum; 
    
    $mix(\cdot)$: function of linear mixture;
    \For{$x$ in loader} 
                \State Calculate $x_1$ and $x_2$ by $aug(x)$
                \State Calculate $x_m$ by $mix(x_1, x_2)$
                \State Extract features $z_1, z_2$ from $f(x_1), f(x_2)$
                \State Calculate $z_m$ by $max(z_1, z_2)$
                \State Extract prediction feature $p_1, p_2, p_m$ from $ h(x_1)$, $h(x_2)$, and $h(f(x_m))$
                \State Compute $\mathcal{L}_{siam}$ in Eq.\ref{eq:losssiam} with $p_1, p_2, z_1$ and $z_2$
                \State Compute $\mathcal{L}_{mix}$ in Eq.\ref{eq:lossmix} with $p_m$ and $z_f$
                \State Compute the final loss with $\mathcal{L}_{siam}$ and $\mathcal{L}_{mix}$
                \State Backward to update the encoder $f$ and predictor $h$
        \EndFor  
  \end{algorithmic}
  \caption{MixSiam}
  \label{algorithm}
\end{algorithm} 

\section{Experiment}
In this section, we evaluate our method on various unsupervised datasets. We first employ the standard linear evaluation protocol to learn representations on ImageNet~\cite{russakovsky2015imagenet}. Then we transfer the learned model to multiple fine-grained image classification datasets. Finally, we analyze our method with ablation studies to demonstrate the effectiveness of each design.
\begin{table}[ht]
\centering
\resizebox{0.47\textwidth}{!}{
\begin{tabular}{lccc}
\hline Method &  Epoch & Top-1 & Top-5 \\
\hline Random &  200 & $5.6$ & $-$ \\
Supervised &  200 & $75.5$ & $-$ \\
\hline 100 epoch training & & & \\
SimSiam~\cite{chen2021exploring} &  100 & $68.1$ & $-$ \\
\textbf{MixSiam (Ours)} &  100 & $\mathbf{69.7}$ & $\mathbf{89.3}$ \\
\hline 200 epoch training & & & \\
LA~\cite{zhuang2019local} &  200 & $60.2$ & $-$ \\
MoCo \cite{he2020momentum}  &  200 & $60.6$ & $-$ \\
CPC-v2~\cite{henaff2020data} &  200 & $63.8$ & $85.3$ \\
CMC~\cite{tian2020contrastive}  &  200 & $64.4$ & $88.2$ \\
MoCo-v2~\cite{chen2020improved}  &  200 & $67.5$ & $-$ \\
MoCHi~\cite{kalantidis2020hard}  &  200 & $68.0$ & \\
CO2 \cite{wei2020co2}  &  200 & $68.0$ & \\
TKC~\cite{feng2021temporal} &  200 & $69.0$ & $88.7$ \\
SimSiam~\cite{chen2021exploring} &  200 & $70.0$ & $-$ \\
\textbf{MixSiam (Ours)} &  200 & $\mathbf{71.1}$ & $\mathbf{90.0}$ \\
\hline 400 epoch training & & & \\
SwAV~\cite{caron2020unsupervised}  &  400 & $70.1$ & $-$ \\
TKC~\cite{feng2021temporal} &  400 & $70.8$ & $89.9$ \\
SimSiam~\cite{chen2021exploring} &  400 & $70.8$ & $-$ \\
\textbf{MixSiam (Ours)} &  400 & $\mathbf{71.9}$ & $\mathbf{90.6}$ \\
\hline 800 \& 1000 epoch training & & & \\
PIRL\cite{misra2020self}  &  800 & $63.6$ & $-$ \\
SimCLR~\cite{chen2020simple} &  1000 & $69.3$ & $89.0$ \\
MoCo-v2 \cite{chen2020improved}  &  800 & $71.1$ & $-$ \\
SimSiam~\cite{chen2021exploring} &  1000 & $71.3$ & $-$ \\
TKC~\cite{feng2021temporal} &  1000 & $72.1$ & $90.6$ \\
\textbf{MixSiam (Ours)} &  800 & $\mathbf{72.3}$ & $\mathbf{90.9}$ \\
\hline
\end{tabular}}
\vspace{0.1cm}
\caption{Comparisons with the state-of-the-art methods under the linear classification protocol on ImageNet. Inspired by~\cite{feng2021temporal}, the Top-1 and Top-5 accuracy with different epochs are reported. The top results at different epochs are highlighted in \textbf{bold}.}
\label{tab:IN_cls}
\end{table}

\subsection{Implementation Details}
\heading{Architecture}
For encoder $f$, we use a convolutional residual network~\cite{he2016deep} with 50 layers as our base parametric encoders followed by a projection multi-layer perceptron (MLP) head. The projection MLP has 3 layers with batch normalization (BN) applied, where the output fc has no ReLU, and the dimension is 2048. For predictor $h$, we adopt BN to its hidden fc layers with the dimension d = 512, while the output fc does not have BN or ReLU. The dimension of the projection MLP's input and output are all set to 2048. 

\textbf{ImageNet} contains 1000 classes, where 1.28 million images are used for training and 50,000 images are used for validation. 

To conduct pre-training on ImageNet, we use the ResNet-50 as our encoder network. We use the SGD optimizer, a cosine decay learning rate schedule\cite{goyal2017accurate} with a base value of 0.05 and weight decay with a value of 0.0001. 128 is set as the default batch size. The hyper-parameter $\lambda_{mix}$ and $\lambda$ are all set to 0.5. We train the encoder network for up to 800 epochs without using either negative samples or a momentum encoder on 4 NVIDIA Tesla V100 GPUs.

For linear evaluation, on top of the frozen features from ResNet’s global average pooling layer, we train a linear classifier to evaluate the representation of our method. The linear classifier training uses base learning rate of 0.02 with a cosine decay schedule for 90 epochs. We set the weight decay to 0, set the momentum to 0.9, and set the batch size to 4096 with a LARS optimizer~\cite{you2017large} on 4 NVIDIA Tesla V100 GPUs. 

\textbf{CIFAR10/100}~\cite{krizhevsky2009learning} both contain 60,000 images, including 50,000 images for training and 10,000 images for test. 

We conduct two settings of experiments. First, we transfer the ImageNet pre-trained model on CIFAR10/100 to show the generalization ability and make a fair comparison with state-of-the-art methods. Specifically, the linear classifier training uses a base learning rate of 0.02 with a cosine decay schedule for 800 epochs, the weight decay is set to 0, the momentum is set to 0.9, and the batch size is set to 512 with a LARS optimizer. 
Second, we train from scratch on CIFAR10 for ablation study. Specifically, ResNet-18 is used as the backbone network, and SGD is adopted as the optimizer with a learning rate of 0.03, the momentum of 0.9, weight decay of 0.0004. Each experiment is repeated 5 times to compute mean and standard derivation.

\textbf{Food101}~\cite{bossard2014food} contains 101 food categories, with 101,000 images. For each class, 250 test images and 750 training images are provided. The training images are not cleaned with some amount of noise. 

For the transfer learning experiment, the encoder is pre-trained on ImageNet, and we transfer them to different datasets for training the linear classifier. The training detail is the same as that of CIFAR10. 

\subsection{Linear Evaluation}

\begin{figure}[t]
   \centering
    \includegraphics[width=0.49\linewidth]{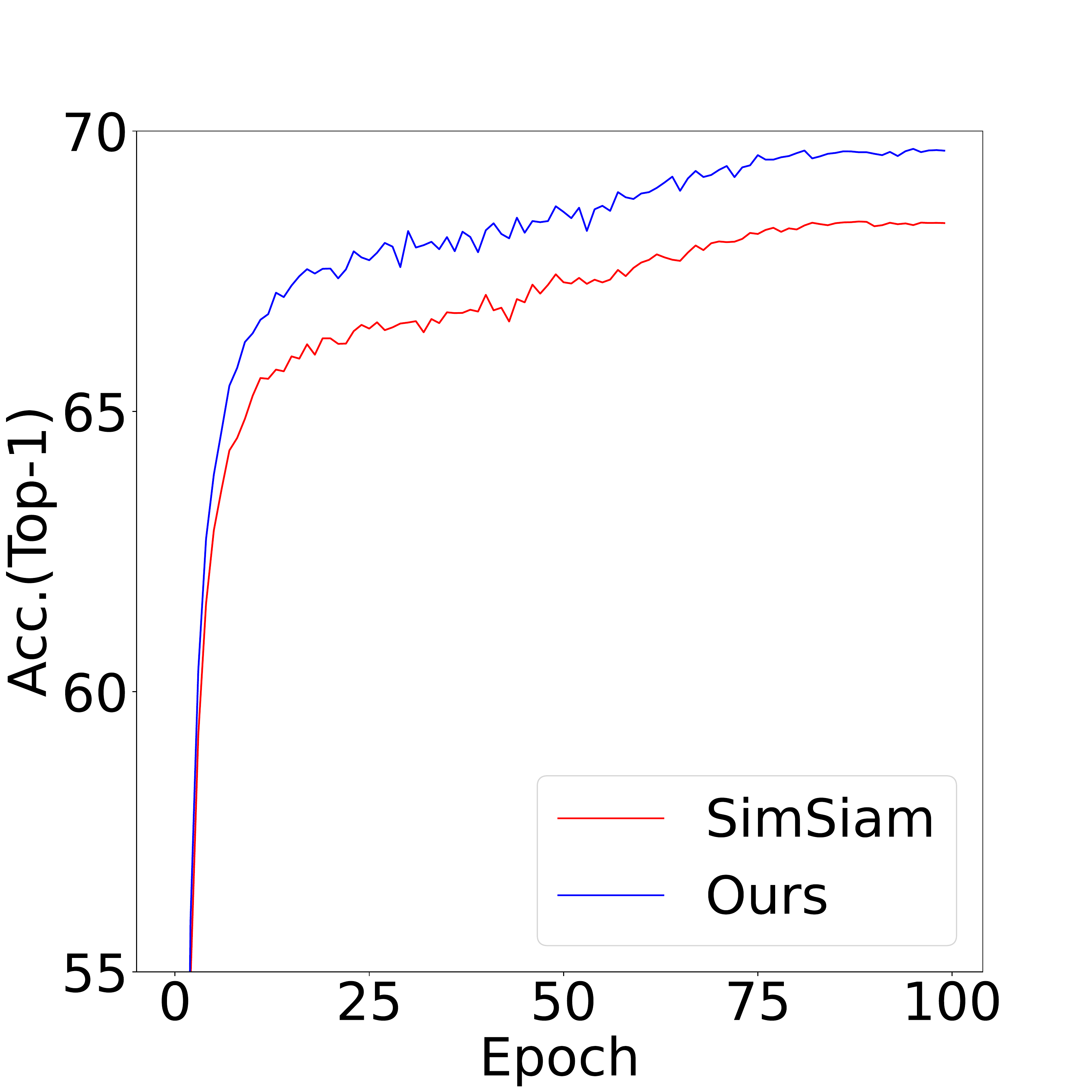}
    \includegraphics[width=0.49\linewidth]{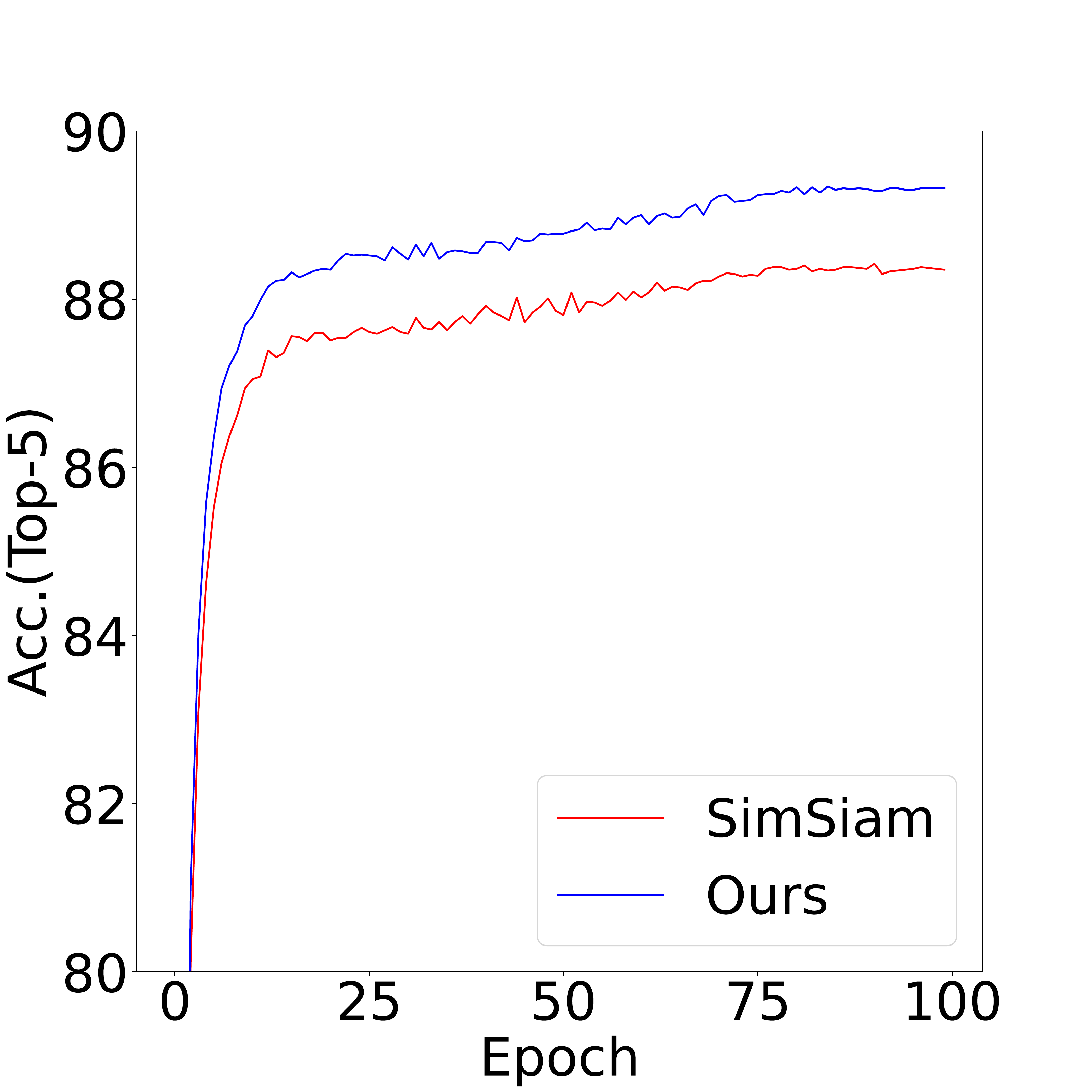}
   \caption{Linear classification accuracy of Top-1 and Top-5 on ImageNet dataset. The results of SimSiam and our method are reported.}
   \label{fig:acc_cmp}
\end{figure}

\heading{Comparison with the SimSiam baseline.} 
The results of our method and SimSiam is shown Fig.~\ref{fig:acc_cmp}. To conduct the experiment, we reproduce SimSiam using the official PyTorch implementation for a fair comparison. We train the model for 100 epochs with a batch size of 128 on 4 NVIDIA Tesla V100 GPUs. We observe that our method continuously outperforms SimSiam as the number of epochs increases. Performances with more training epochs are reported in Table~\ref{tab:IN_cls}, where we adopt the SimSiam result from the original paper. We observe that our method improves SimSiam by 1.6\% and 1.1\% on 100 and 200 epochs, respectively. The experimental results demonstrate the effectiveness of the proposed strategy of adopting the mixed samples. It is obvious that learning from the unseen mixed images and the constraint with the discriminative representation contributes to the simple siamese structure. 

\heading{Comparison with the state-of-the-arts.} We report top-1 and top-5 accuracy of our method and the state-of-the-art methods in Table~\ref{tab:IN_cls}. Performances at different epochs from 100 to 800 are listed. Compared to previous works on different pretext tasks, the performance of our method are also superior, which indicates that learning from unseen mixed image with feature aggregation can provide representation with better discriminative ability.

\begin{table}[t]
\centering
\small
\resizebox{0.47\textwidth}{!}{
\begin{tabular}{lccc}
\hline Method & Food101 & CIFAR10 & CIFAR100   \\
\hline 
Supervised-IN & $72.3$ & $93.6$ & $78.3$  \\
\hline
InsDis \cite{wu2018unsupervised}& 63.4 &80.3 &59.9\\
PIRL \cite{misra2020self}  &  64.7 & 82.55 &61.3\\
MoCo-v2 \cite{chen2020improved}&68.9&92.3&74.9\\
SimCLR~\cite{chen2020simple} & $68.4$ & $90.6$ & $71.6$  \\
InfoMin \cite{tian2020makes} & 69.5& 91.5&73.4\\
\textbf{MixSiam (Ours)} & $\textbf{76.5}$ & $\textbf{93.0}$ & $\textbf{77.3}$  \\
\hline
\end{tabular}}
\caption{Transfer learning results in linear evaluation on Food101, CIFAR10 and CIFAR100. The performances of the supervised baseline, the-state-of-the-art methods as well as our method are reported. The top results are highlighted in \textbf{bold}.}
\label{tab:tran_cls}
\end{table}
\subsection{Transfer Learning} We discuss the effectiveness of the representations learned for transfer learning on the fine-grained image classification task. As shown in Table~\ref{tab:tran_cls}, for the state-of-the-art methods and our method, the top-1 accuracy is reported on CIFAR10/100 and Food101. Note that our method surpasses the model trained on supervised ImageNet on Food101, while providing only slightly worse performance on CIFAR10 and CIFAR100. Our method also outperforms other state-of-the-art methods such as SimCLR~\cite{chen2020simple} on all benchmarks. The result demonstrates that our learned representation is capable when transferred across image domains. 

\subsection{Ablation studies}
\textbf{The impact of feature aggregation.} The performances with different feature aggregation strategies are shown in Table~\ref{tab:ablation}. We observe that the performance using the max operation is higher than the other two strategies with 0.64 and 0.49 points of improvement, respectively. The reason is that the max operation considers the features with the highest scores, where the most representative information is maintained, \textit{i.e.}, discriminative representation. During training, our network forces the prediction of the mixed hard samples to express the discriminative representation, where the hard samples are exhausted studied, which enhances the robustness of the model and boosts the performance. 

We also observe that when conducting the average operation, the performance is slightly lower than without any aggregation with 0.15 points. This is reasonable. The explanation is that the average operation takes the whole image into count without considering the the most informative feature, which could further result in a less discriminative aggregated representation.

\textbf{The impact of image mixture.} We show the performance with/without image mixture in Table~\ref{tab:ablation}. We observe that by adopting image mixture, the performance improves 2.64 points. The impressive improvement demonstrates the effectiveness of the mix-based feature learning strategy. Learning with the generated mixed samples enlarges the training distribution, where the hard samples help to enhance the robustness of the model. Moreover, since the other two inputs are random augmented views with potentially large mutual information, the model relied on only augmented pairs could be difficult to deal with the large intra-class variation. Learning from these hard virtual samples improves the discrimination ability.

\begin{table}[t] 
\centering
\setlength{\tabcolsep}{7mm}{
\begin{tabular}{c|c|c}
\hline Component & Method & Acc.\\
\hline \multirow{3}*{\shortstack{Feature\\Aggregation}}&Maximum& $\textbf{93.35}$\\
~&Average& $92.71$ \\
~&W/O & $92.86$ \\
\hline \multirow{2}*{\shortstack{Image\\Mixture}}&Mixture& $\textbf{93.35}$ \\
~&W/O & $90.71$ \\
\hline
\end{tabular}}
\caption{Method analysis on CIFAR10. The performance of using different feature aggregation strategies and image mixture strategies are reported. When without feature aggregation, the feature from either branch is adopted to replace the aggregated feature. When without image mixture, one of the augmented images is selected randomly to replace the mixed image. The top results are highlighted in \textbf{bold}.}
\label{tab:ablation}
\end{table}
\begin{figure}[t]
   \centering
    \includegraphics[width=0.7\linewidth]{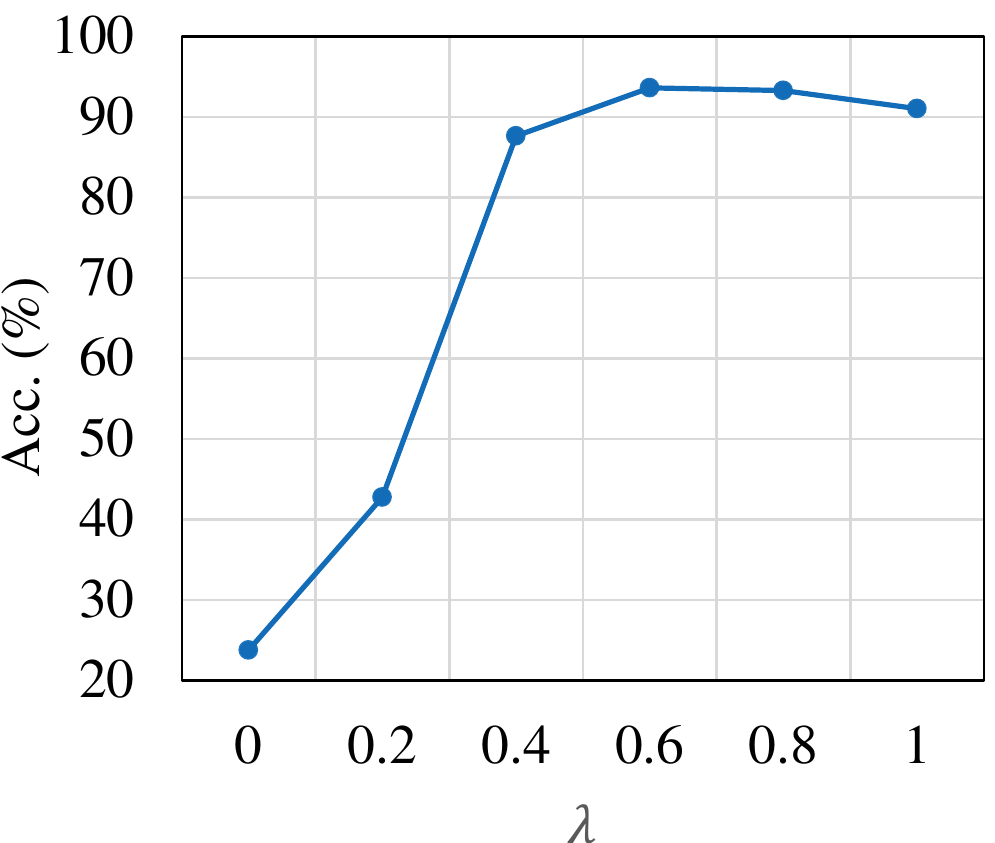}
   \caption{Performance curve with different values of parameter $\lambda$ on CIFAR10, where $\lambda$ balances the two kinds of losses.}
   \label{fig:lambda}
\end{figure}
\begin{figure*}[ht]
   \centering
    \includegraphics[width=0.95\linewidth]{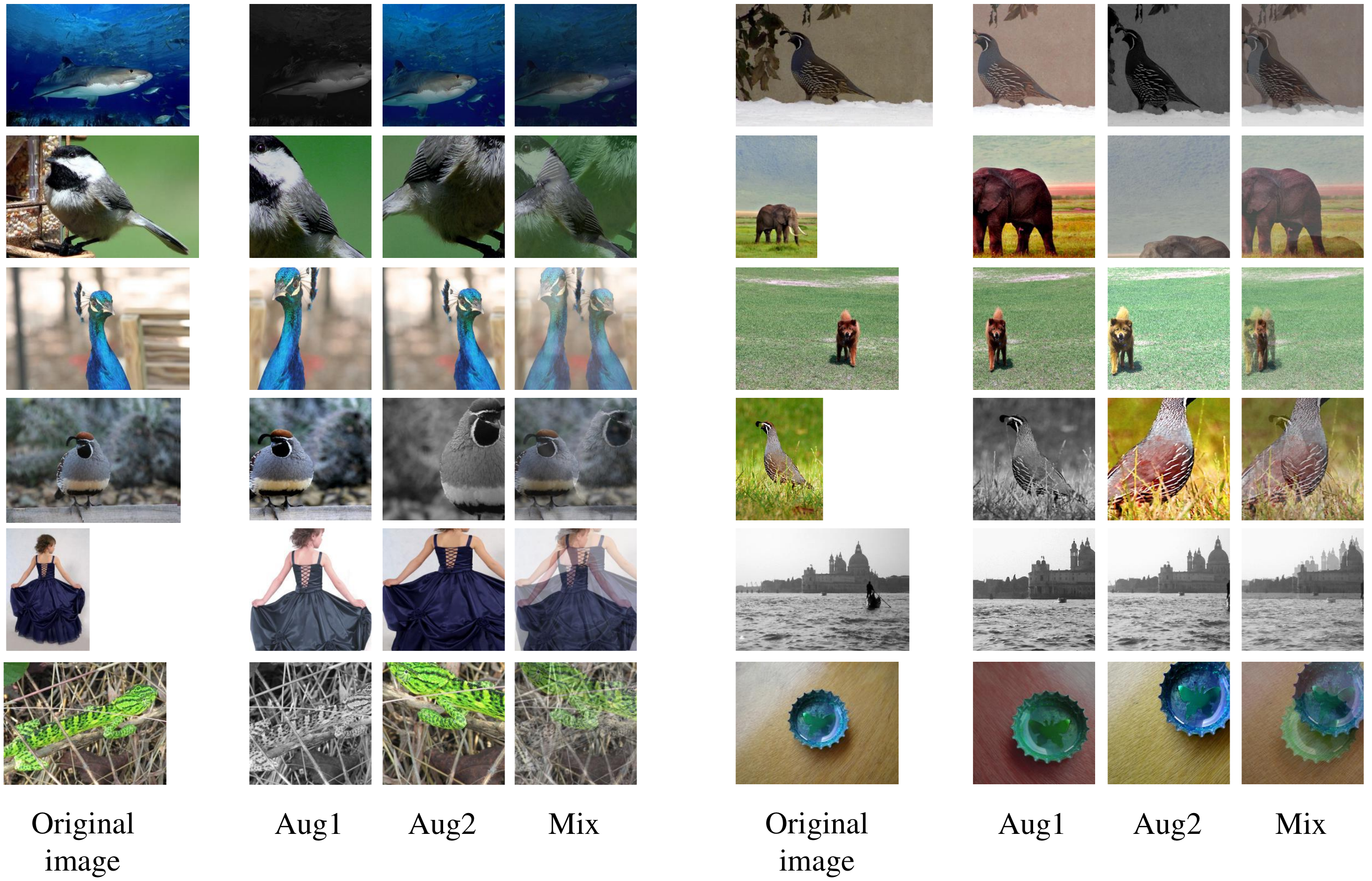}
   \caption{Visualization of the input images. The original images, the two random augmented images (denoted by `Aug1' and `Aug2') and the mixed image (denoted by `Mix') are shown.}
   \label{fig:visual}
\end{figure*}

\textbf{The impact of the hyper-parameter $\lambda$}. In Eq.~\ref{eq:final}, the hyper-parameter $\lambda$ balances the two losses of the easy positive pairs and the hard mixed pairs. We vary $\lambda$ from 0 to 1 and report the corresponding performance in Fig.~\ref{fig:lambda}. 
When $\lambda$ equals 0, the loss with mixed image will dominate, where we obtain an accuracy of 23.76\%. The poor performance demonstrates that learning with only mixed samples and the discriminative representation is not enough. The reason is that the mixed training samples consist of huge variance, which is unusual in the real world (the test set). Training with only mixed images will reduce the network's perception of a large number of easy samples. 
When $\lambda$ equals 1, only the relationship between easy augmented pairs is considered and the method will be the same as SimSiam~\cite{chen2021exploring}. We observe that when $\lambda$ increases from 0 to 0.6, the performance continues increasing and becomes higher than that of SimSiam. This demonstrates that, with the help of harder samples and the stronger constraint, the robustness of the model is improved and more discriminative representation is learned. Finally, a value of 0.5 is selected for the hyper-parameter $\lambda$.

\subsection{Visualization}
We visualize the original training sample, the randomly augmented two views and the mixed image in Fig.~\ref{fig:visual}, where multiple examples are shown. Note that, we adopt the same set of data augmentation strategies as in SimSiam~\cite{chen2021exploring}, including geometric augmentation, color augmentation and blurring augmentation. Concretely, the procedure of data augmentation is described as the PyTorch\cite{paszke2019pytorch} notations: we first use RandomResizedCrop to select a random patch of the images and resize the patch to 224 x 224 with RandomHorizontalFlip; then we adjust brightness, contrast, saturation and hue of 0.8 probability as a color distortion, and perform grayscale conversion of 0.2 probability with RandomGrayscale; finally, we apply Gaussian blur to the patches. 

As shown in Fig.~\ref{fig:visual}, the augmented views, though they go through crop and color distortion, are easier to be recognized with their clear edges and contexts. For the mixed image, by averaging the two views, the image is visually blurred and the overlapped two target objects introduce further variance. This makes the task more challenging to predict the invariant discriminative representation for the hard mixed samples. In turn, the model benefits from the hard sample learning process. 

\section{Conclusion}
In this paper, we propose MixSiam, a mixture-based approach upon the siamese contrastive. 
By investigating the state-of-the-art siamese-based method, we find it difficult to deal with large intra-class variations as it only learns from augmented views with little variance. To this end, we present a mix-based feature learning strategy upon the siamese structure. On the one hand, the hard virtual samples generated by mixup are introduced to enhance the robustness. On the other hand, we capture the discriminative representation of an instance from its corresponding augmented views. By learning to minimize the similarity between the hard virtual sample and the discriminative representation, the model learns to be invariant to the hard samples. Experiments are conducted on standard ImageNet and several small datasets, demonstrating our superiority comparing with the state-of-the-art methods. The consistent improvement over the siamese baseline further verifies the effectiveness of our method. In the future, the augmentation strategies will be investigated to select the appropriate ones for generating the mixed images. More experiments will be conducted on more downstream tasks such as object detection and object segmentation to further validate the generalization ability of the proposed model. 
{
\footnotesize
\bibliography{aaai22.bib}
}
\end{document}